\documentclass[10pt,twocolumn,letterpaper]{article}

\usepackage{cvpr}              % To produce the CAMERA-READY version
\definecolor{cvprblue}{rgb}{0.21,0.49,0.74}
\usepackage[pagebackref,breaklinks,colorlinks,allcolors=cvprblue]{hyperref}
\usepackage{booktabs}
\usepackage{multirow}
\usepackage{amsmath}
\usepackage{pifont}
\newcommand{\cmark}{\ding{51}}
\newcommand{\xmark}{\ding{55}}
\usepackage{float}
\usepackage{caption}
\usepackage[accsupp]{axessibility}
\def\paperID{} % *** Enter the Paper ID here
\def\confName{CVPR}
\def\confYear{2026}

\title{Cross-Axis Feature Fusion with Joint-Wise Motion Difference Prediction for Text-Based 3D Human Motion Editing}

\author{Gyojin Han\\
School of Electrical Engineering, KAIST\\
{\tt\small hangj0820@kaist.ac.kr}
\and
Junmo Kim\\
School of Electrical Engineering, KAIST\\
{\tt\small junmo.kim@kaist.ac.kr}
}

\begin{document}
\maketitle
\begin{abstract}
We address text-based 3D human motion editing, where the goal is to preserve the style and structure of a source motion while applying edits described in natural language. The release of the MotionFix dataset has spurred active research into training-based diffusion models that directly generate an edited motion from a source motion and a text instruction. While previous works have focused primarily on learning when an edit should occur temporally, our goal is to create a model that understands not only this temporal aspect but also which specific joints are responsible for the change. Targeting this, we propose a novel architecture and a complementary auxiliary task to aid its training. Our architecture consists of two axis-anchored transformers, which extract distinct features along the joint and time dimensions respectively, and a cross-axis fusion block that integrates these representations. We further introduce an auxiliary task that trains the joint-anchored transformer to regress the Soft-DTW distance between source and target joint rotations. This objective teaches the module to understand which joints to modify and which to preserve. Through comprehensive experiments on the MotionFix dataset, we demonstrate that our method significantly improves semantic alignment with both the text instruction and the source motion, as well as the overall fidelity of the generated motion, achieving state-of-the-art results.
\end{abstract}    
\section{Introduction}
\label{sec:intro}

3D human motion generation~\cite{Action2Motion2020,ACTOR2021,ActFormer2022, MDM2023,MotionDiffuse2022,MLD2023,InterDiff2023,LA_GMDM_ECCV2024,MARDM_CVPR2025,GUESS2024,EnergyMoGen2024,DART2024,MotionCLR2024} and editing~\cite{MotionFix2024,SimMotionEdit2025,IterativeME2024,MotionCLR2024,DynBlend2025} play central roles in content creation for animation, film, and games. While generation enables rapid prototyping of new sequences from textual prompts, editing is particularly practical when teams must adapt an existing take without discarding its strengths. For example, one may wish to preserve an animator’s style or a captured actor’s trajectory while making targeted adjustments late in production, such as changing which arm lifts, shifting the timing of a turn, or slightly exaggerating a jump. In these scenarios, text-based editing offers a concise interface for specifying the desired modifications while maintaining continuity with the source sequence.

Early approaches~\cite{10.1145/253284.253321, 10.1145/311535.311539, 10.1145/364338.364400, 10.1145/54852.378507,10.1145/3386569.3392469,10.1145/2897824.2925975, PoseFix2023,10.1145/3550454.3555454} to motion editing often relied on task-specific modules for stylization, pose editing, or in-betweening and non-textual inputs like spatial or temporal constraints. While these methods improved controllability, they lacked the flexibility of text-based instructions and offered limited joint-level specificity. The advent of diffusion models~\cite{DDPM, song2021denoising, NEURIPS2022_260a14ac, ho2022classifier, DiT, Hur_2024_WACV} introduced powerful text-to-motion generation capabilities, yet most systems~\cite{MDM2023,MotionDiffuse2022,MLD2023,InterDiff2023,LA_GMDM_ECCV2024,MARDM_CVPR2025,GUESS2024,EnergyMoGen2024,DART2024,MotionCLR2024} focused on synthesizing motion from scratch rather than minimally modifying an existing sequence. 

A significant shift occurred with the release of the MotionFix dataset~\cite{MotionFix2024}. By providing triplets of text instruction, source motion, and target motion, it enabled a supervised setting for the editing task. The accompanying baseline, TMED~\cite{MotionFix2024}, operates in this setting by concatenating a CLIP-based text embedding~\cite{CLIP} with a shallow projection of the source motion, feeding the resulting tokens directly to a diffusion transformer as a conditioning signal. Building on this, SimMotionEdit~\cite{SimMotionEdit2025} proposed a condition transformer to fuse text and motion features before they are passed to the diffusion transformer (DiT)~\cite{DiT}. SimMotionEdit also introduced an auxiliary motion similarity prediction task to help the model learn when in the sequence changes should occur. However, the SimMotionEdit architecture aggregates features along the joint dimension for each time step, which constrains the extraction of disentangled joint-level information. Consequently, its auxiliary supervision is also restricted to the frame level (predicting frame-wise similarity). While this provides temporal guidance, it offers limited insight into which specific joints are responsible for the change. Therefore, the encoder's understanding of joint-level control remains under-specified, limiting the semantic alignment between the condition and the resulting edits.

To address this limitation, we introduce a new architecture composed of two axis-anchored transformers and a cross-axis fusion block, together with a complementary joint-wise auxiliary task. Our architecture is designed to separately model the two primary axes of motion data. First, a joint-anchored transformer aggregates the global trajectory for each joint across the entire sequence. Concurrently, a time-anchored transformer captures the full-body pose characteristics for each individual frame. These two distinct representations are then integrated by a cross-axis fusion block, which allows frame-wise features to attend to the joint-wise context. The resulting fused features are used to condition the diffusion transformer for generation.
To further guide the joint-anchored module, our joint-wise auxiliary task regresses the per-joint Soft-DTW distance between the source and target rotation trajectories, using the features from the joint-anchored transformer. This auxiliary objective thereby trains the joint-anchored module to comprehend which specific joints must be modified and which must be preserved to accurately reflect the text instruction.
We validate the approach through extensive experiments and ablations on the MotionFix dataset, demonstrating consistent improvements over prior methods.

We can summarize the contributions of this work as follows:
\begin{itemize}
\item We propose a novel conditioning architecture featuring axis-anchored transformers to separately model the spatial and temporal axes of motion. These are integrated by a cross-axis fusion block to provide joint-aware and temporally-aware conditioning for the diffusion model.

\item We introduce a joint-wise motion difference prediction objective, an auxiliary task that uses Soft-DTW over rotation trajectories. This provides timing-robust, joint-specific supervision, guiding the model to learn which joints to modify and how much they should change.

\item We conduct comprehensive validation on the MotionFix dataset, including extensive ablation studies that analyze the efficacy of our proposed architecture, supervision design, and fusion strategy.
\end{itemize}

\section{Related Works}
\label{sec:related_works}

\subsection{3D Human Motion Generation}
Early approaches to 3D human motion generation relied on recurrent models, variational autoencoders, and generative adversarial networks~\cite{Action2Motion2020,ACTOR2021,ActFormer2022}.
With the emergence of diffusion models and rapid gains in their generative capability, diffusion-based text-to-motion methods~\cite{MDM2023,MotionDiffuse2022,MLD2023,InterDiff2023,LA_GMDM_ECCV2024,MARDM_CVPR2025,GUESS2024,EnergyMoGen2024,DART2024,MotionCLR2024} trained on large-scale paired text–motion datasets~\cite{HumanML3D2022,KITML2016,BABEL2021} have become the dominant paradigm.
This shift has delivered steady improvements in fidelity and diversity across benchmarks.
Modern pipelines typically predict sequences of parametric human body poses (e.g., SMPL parameters) and often incorporate auxiliary controls such as trajectories, foot contacts, or end-effector constraints~\cite{SMPL2015,OmniControl2024,InterDiff2023}. 
Recent advances further extend conditioning beyond text to shape- or scene-aware signals~\cite{TapMo2023,SMD2024}, and explore online or interactive control regimes for real-time applications~\cite{DART2024}. 
Collectively, these works establish a progression from autoregressive and latent-variable generators to diffusion-driven models that integrate multimodal conditioning and explicit control signals for scalable, high-quality motion synthesis.

\subsection{3D Human Motion Editing}
Building on progress in generation, research on human motion editing has advanced rapidly, aiming to transform an input motion according to user intent while preserving identity- or style-consistent attributes~\cite{MotionFix2024,SimMotionEdit2025,IterativeME2024,MotionCLR2024,DynBlend2025}. 
Earlier editing methods typically imposed spatial or temporal constraints~\cite{10.1145/253284.253321, 10.1145/311535.311539, 10.1145/364338.364400, 10.1145/54852.378507}, or decomposed authoring into stylization, pose editing, and in-betweening subproblems~\cite{10.1145/3386569.3392469,10.1145/2897824.2925975, PoseFix2023,10.1145/3550454.3555454}; these designs improved controllability but required non-text inputs and offered limited joint-level specificity. 
Subsequent diffusion-based editing introduced masked inpainting~\cite{KimInpaintEdit2024}, attention manipulation~\cite{MotionCLR2024}, and keyframe-guided heuristics~\cite{IterativeME2024}, which enhanced directability yet continued to rely on handcrafted interfaces. 
Text-driven editing then emerged either at the pose level~\cite{PoseFix2023} or through indirect prompting via large language models~\cite{FineMoGen,CoMo}, but these approaches still struggled to realize fine-grained, joint-aware edits across full-body sequences. 
The MotionFix dataset~\cite{MotionFix2024} addresses this gap by providing instruction--source--target triplets that enable direct supervision for text-based editing, catalyzing conditional diffusion methods that leverage both the source motion and textual instructions. 
Within this trajectory, TMED~\cite{MotionFix2024} operationalizes conditional diffusion on MotionFix to align edits with natural-language instructions, while SimMotionEdit~\cite{SimMotionEdit2025} integrates motion-similarity prediction and a new transformer design to strengthen content preservation and instruction faithfulness during editing. 
Going beyond prior methods, our approach explicitly specifies which joints should change and how by introducing joint-anchored and time-anchored transformers with a cross-axis fusion block and a joint-wise Soft-DTW auxiliary objective, enabling instruction-faithful, joint-aware edits.

\subsection{Diffusion Models}
Diffusion models~\cite{DDPM} were first popularized in image synthesis, where iterative denoising from Gaussian noise—together with improved parameterizations, samplers, and guidance—yielded large gains in fidelity and controllability~\cite{NEURIPS2021_49ad23d1, Rombach_2022_CVPR}. Key developments include score-based/denoising diffusion formulations, accelerated samplers, classifier-free guidance, and transformer-based backbones that scale well with data and compute~\cite{song2021denoising, NEURIPS2022_260a14ac, ho2022classifier, DiT}. The same principles were subsequently extended beyond images to video and 3D content, demonstrating that diffusion in an appropriate latent or parameter space can model complex spatio–temporal structure~\cite{NEURIPS2022_39235c56, ho2022imagen, shi2024mvdream, 10.1007/978-3-031-73220-1_11}.
\section{Method}

\begin{figure*}[t]
  \centering
    \includegraphics[width=1.0\linewidth]{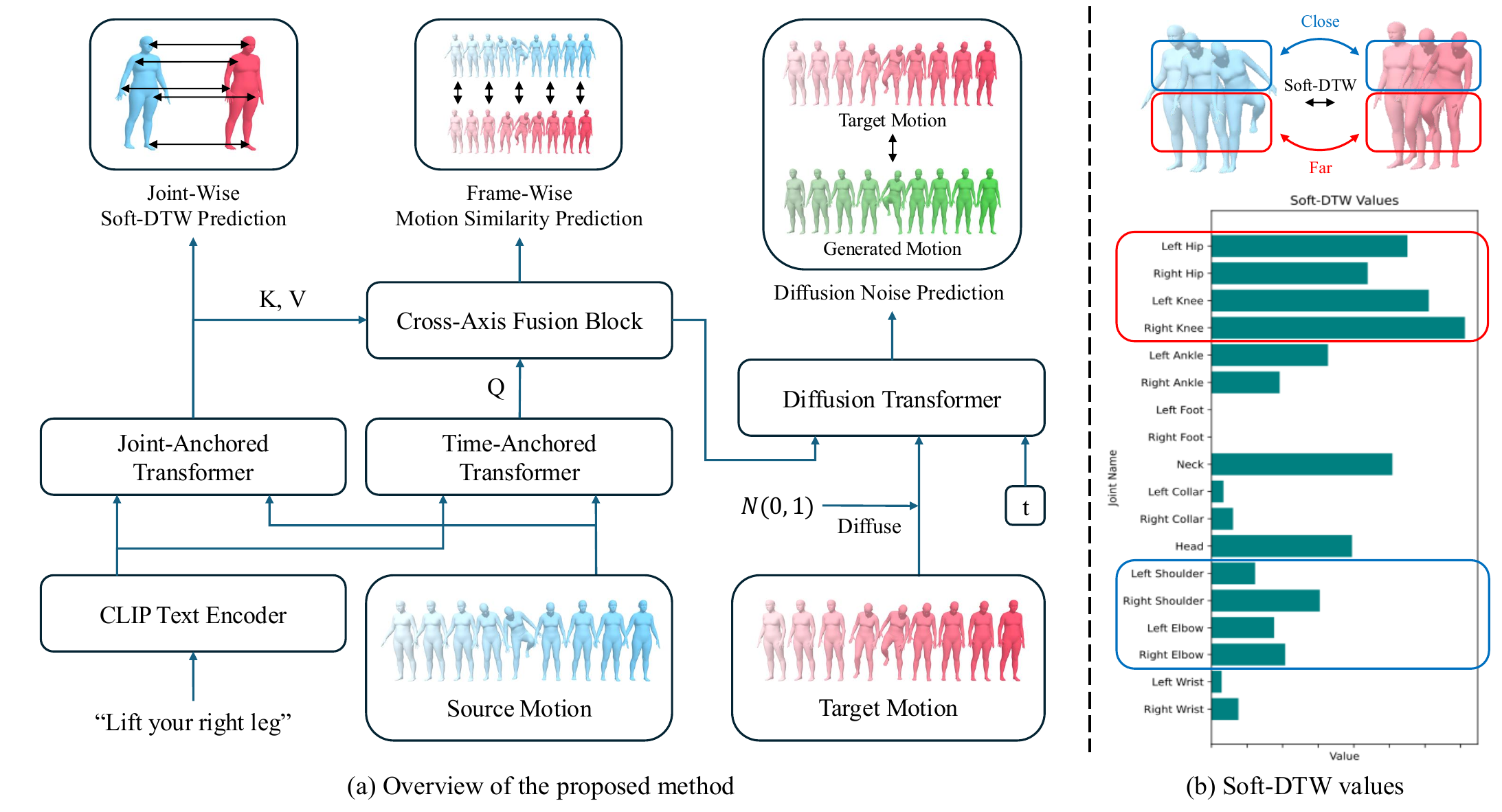}
      \caption{Overview of the proposed approach and joint-wise supervision.
\textbf{(a)} A joint-anchored transformer and a time-anchored transformer produce enhanced features, $h_{\text{joint}}$ and $h_{\text{time}}$. A cross-axis fusion block performs multi-head attention with these and the fused representation conditions a DiT that predicts denoising updates to generate the edited motion from noise, given the source motion and the text. Two auxiliary heads supervise the encoders by frame-wise motion similarity and joint-wise Soft-DTW prediction.
\textbf{(b)} Example Soft-DTW values between the source and target rotation trajectories for some joints. We observe that joints exhibiting larger motion changes yield higher Soft-DTW distances than those with smaller changes.}

  \label{fig:method}
\end{figure*}

In this section, we present the architecture and training methodology of our proposed text-based 3D human motion editing model. The model consists of a joint-anchored transformer, a time-anchored transformer, and a cross-axis fusion block for encoding features from the source motion and the text instruction. For the motion generation process, we employ a DiT. For the effective training of this architecture, we introduce a novel auxiliary task, termed joint-wise motion distance prediction. An overview of our method is illustrated in Fig.~\ref{fig:method}, and a detailed elaboration of each component will be provided in the following subsections.

\subsection{Diffusion Model for Human Motion Editing}

The recent release of the MotionFix dataset~\cite{MotionFix2024}, which provides triplets of text instructions, source motions, and target motions, has facilitated recent progress in text-based 3D human motion editing. This dataset allows direct training of diffusion models for this task, and our work focuses on this setting.
In this setting, the source motion $S = \{s^i\}_{i=1}^N$ and target motion $T = \{t^i\}_{i=1}^N$ are represented as sequences of human body poses, where $N$ denotes the number of frames in each sequence. Following a previous work~\cite{MotionFix2024, SimMotionEdit2025}, we represent each source pose $s^i \in \mathbb{R}^K$ and target pose $t^i \in \mathbb{R}^K$ as a vector with $K=207$ dimensions. Specifically, this 207-dimensional vector comprises: 126 (21 $\times$ 6) 6D rotation parameters for 21 joints of the SMPL model (excluding the left hand, right hand, and root joint); 66 (22 $\times$ 3) position parameters for 22 joints (all SMPL joints excluding the left and right hands); 3 global translation parameters; and 12 global orientation parameters.

Our goal is to generate an edited motion $M = \{m^i\}_{i=1}^N$ from an input source motion $S$ and a text prompt $c$ using a diffusion model. For this, the diffusion model $g$ is trained to minimize the noise-prediction loss:
\begin{equation}
\mathcal{L}_{\text{diff}}
= \mathbb{E}_{\tau,\epsilon}\left[
\left\| g\left(T_\tau; f(S,c), \tau\right) - \epsilon \right\|_2^2
\right],
\label{eq:diffusion-eps}
\end{equation}
where $\tau$ is a diffusion timestep sampled uniformly from $\{1,\dots,\mathcal{T}\}$, $\epsilon \sim \mathcal{N}(0,I)$, and $f$ is the feature encoder for the source motion $S$ and text prompt $c$.
More specifically, the noisy target motion $T_\tau$ is constructed by the forward diffusion process
$T_\tau = \sqrt{\bar{\alpha}_\tau} T + \sqrt{1-\bar{\alpha}_\tau} \epsilon$.
where $\bar{\alpha}_\tau = \prod_{s=1}^{\tau} \alpha_s$ and $\alpha_s = 1-\beta_s$, with a variance schedule $\{\beta_s\}_{s=1}^{\mathcal{T}}$.
At inference time, we synthesize the edited motion by running the DDPM reverse process from pure noise. We initialize $M_{\mathcal{T}} \sim \mathcal{N}(0,I)$ and, for $\tau=\mathcal{T},\dots,2$, we compute
\begin{equation}
\hat{\epsilon}_\tau= g\left(M_\tau; f(S,c),\tau\right),
\end{equation}
and sample
\begin{equation}
M_{\tau-1}=\frac{1}{\sqrt{\alpha_\tau}}\left(
M_\tau - \frac{1-\alpha_\tau}{\sqrt{1-\bar{\alpha}_\tau}}\hat{\epsilon}_\tau
\right)+
\sqrt{\tilde{\beta}_\tau}z_\tau,
\end{equation}
where $z_\tau \sim \mathcal{N}(0,I)$ and $\tilde{\beta}_\tau = \dfrac{1-\bar{\alpha}_{\tau-1}}{1-\bar{\alpha}_\tau} \beta_\tau$. After iterating to $\tau=1$, we take the final edited motion as $M = M_0$.

\subsection{Axis-Anchored Transformers and Cross-Axis Fusion Block}
The generated motion $M$ should preserve the structural and stylistic properties of the source motion $S$ while integrating the edit specified by the text instruction $c$. To this end, the feature encoder $f$ must (1) establish fine-grained cross-modal alignment between $S$ and $c$, (2) disentangle edit-relevant factors from content that should remain unchanged, and (3) deliver the resulting conditioning signals to the diffusion model so that denoising applies the intended edit without drifting from $S$. We now examine the design of $f$, the conditioning pathway, and the associated training objectives in detail.

In TMED~\cite{MotionFix2024}, the feature encoder, $f_\text{tmed}$, processes the text instruction using a CLIP text encoder and the source motion using a single linear projection layer. The resulting output embeddings are then concatenated. Notably, no explicit fusion occurs between the text and source motion embeddings within the encoder itself. Instead, information exchange is deferred until these embeddings are fed as tokens into the DiT, where fusion occurs via self-attention mechanisms between the text embedding, the source motion embedding, and the noised motion. 

SimMotionEdit~\cite{SimMotionEdit2025} introduced a condition transformer as a feature encoder $f_\text{sim}$, to process both text and source motion embeddings, attempting feature fusion prior to the DiT. Specifically, SimMotionEdit projects the $K$ dimensions of each source pose into a latent dimension $D$. It then fuses these projected features with the text embedding to obtain an enhanced text feature $h_\text{text} \in \mathbb{R}^{1, D}$ and an enhanced motion feature $h_\text{motion} \in \mathbb{R}^{N, D}$, where pose characteristics are aggregated per frame. Furthermore, SimMotionEdit trained its condition encoder with an auxiliary task: predicting the frame-wise motion similarity between the source and target motions. This was designed to help the model understand when (at which temporal step) the motion modification should occur. By feeding these enhanced $h_\text{text}$ and $h_\text{motion}$ as conditioning signals into the DiT, SimMotionEdit achieved results that demonstrated better alignment with the text instruction compared to TMED. 

However, while the SimMotionEdit approach may help in understanding when in the sequence a change should occur, its reliance on predicting a global, frame-wise similarity metric provides limited information regarding where in the body the edit should be applied. The model is trained to predict the overall similarity difference per frame, which does not help it learn which specific joints should undergo significant modification versus which joints should remain consistent with the source motion. For instance, if the source motion at frame $n$ involves lifting the left leg, and the text instruction is "Lift your right leg," SimMotionEdit's architecture might help identify that a change is needed at frame $n$. However, it provides no explicit guidance for the model to understand that the hip and knee joints require substantial modification, while the upper body joints should remain relatively unchanged.

We claim that for the text-based motion editing task, capturing the global characteristics of each joint across the entire sequence is as important as understanding the temporal characteristics of the pose at each frame. Therefore, we attempt to represent these two axes distinctly using two anchored transformers. Both the joint-anchored transformer and the time-anchored transformer take the source motion $S \in \mathbb{R}^{N \times K}$ and the text instruction $c$ as input. They output a joint-wise feature $h_\text{joint} \in \mathbb{R}^{K, D}$ and a frame-wise feature $h_\text{time} \in \mathbb{R}^{N, D}$, respectively. Following this, we introduce a cross-axis fusion block—implemented as a multi-head attention block—to integrate the information from both features. This block receives $h_\text{time}$ as the Query and $h_\text{joint}$ as both the Key and Value. The resulting fused feature, $h_\text{fusion}$, is then concatenated with the noised motion input to the DiT, thereby conditioning the diffusion process on both the source motion and the text instruction. Through this architecture comprising the joint-anchored transformer, the time-anchored transformer, and the cross-axis fusion block, we have designed a feature encoder $f_\text{fusion}$ that comprehends not only when the motion should be altered but also which specific joints require modification. The architecture of our model is illustrated in Fig.~\ref{fig:method} (a).

\subsection{Joint-Wise Motion Difference Prediction}

In addition to the above architecture, we introduce an auxiliary task that encourages the joint-anchored transformer to learn joint-specific dynamics as intended. This task regresses, for each rotation channel, a scalar measure of the motion difference between the source motion $S$ and the target motion $T$. Because joint motion is best represented by rotation parameters, we first extract from the joint-wise feature $h_\text{\text{joint}}$ the sub-tensor corresponding to the 21 SMPL joints with 6D rotations, denoted $h'_{\text{joint}} \in \mathbb{R}^{K'\times D}$, where $K'=126$. The regression head $\varphi_{\text{reg}}$ takes $h'_{\text{joint}}$ and outputs a per-channel score that should reflect how much the rotation trajectory changes from $S$ to $T$ while being insensitive to timing shifts. Because $h'_{\text{joint}}$ aggregates information over time and many text edits change the start time or speed without altering motion morphology, a frame-by-frame loss would over-penalize harmless timing shifts. We therefore need a distance that is robust to local time warps and compares motions by trajectory shape rather than absolute timing. To provide such supervision, we use Soft-Dynamic Time Warping (Soft-DTW) as the ground-truth target.

Specifically, let $S',T'\in\mathbb{R}^{K'\times N}$ be the rotation-only versions of $S$ and $T$. For each rotation parameter $j\in\{1,\dots,K'\}$, define the time series
$S'_j=(s_{j,1},\dots,s_{j,N})$ and $T'_j=(t_{j,1},\dots,t_{j,N})$. The regression head predicts a scalar score per channel,
\begin{equation}
\hat{d}_j \;=\; \bigl[\varphi_{\text{reg}}(h'_{\text{joint}})\bigr]_j \in \mathbb{R},
\qquad j=1,\ldots,K'.
\end{equation}
We construct the ground-truth score using dynamic time warping. Classical DTW compares two sequences up to local time shifts by minimizing the accumulated pointwise cost over monotone alignment paths. With $d(x_n,y_m)=\|x_n-y_m\|_2^2$,
\begin{equation}
\mathrm{DTW}(x,y)\;=\;\min_{\pi\in\mathcal A}\ \sum_{(n,m)\in\pi} d(x_n,y_m),
\end{equation}
where $\mathcal A$ is the set of admissible monotone warping paths. Soft-DTW replaces the hard minimum with a smooth soft-min (log-sum-exp) controlled by temperature $\gamma>0$,
\begin{equation}
\mathrm{SoftDTW}_\gamma(x,y)\;=\;\operatorname{softmin}^{(\gamma)}_{\pi\in\mathcal A}\ \sum_{(n,m)\in\pi} d(x_n,y_m),
\end{equation}
which preserves DTW’s invariance to local time warps while being fully differentiable. We adopt Soft-DTW in practice and use available GPU-accelerated implementations for efficient training.

Our auxiliary supervision targets the Soft-DTW distance between the source and target rotation trajectories for each channel,
\begin{equation}
d_j \;=\; \mathrm{SoftDTW}_\gamma\!\bigl(S'_j,\;T'_j\bigr)\in\mathbb{R},
\end{equation}
and trains the regression head to match these values from $h'_{\text{joint}}$ via a mean-squared loss,
\begin{equation}
\mathcal{L}_{\text{aux}}
\;=\;
\frac{1}{K'}\sum_{j=1}^{K'}\bigl(\hat{d}_j - d_j\bigr)^2.
\end{equation}
This objective guides the joint-anchored transformer to encode joint-wise motion differences that capture similarity in trajectory shape while remaining robust to differences in start time and duration, complementing the main diffusion objective. As an example, we visualize the Soft-DTW measurements for a sample as a graph in Fig.~\ref{fig:method} (b).

\section{Experiments}

\begin{figure*}[t]
  \centering
\includegraphics[width=1.0\linewidth]{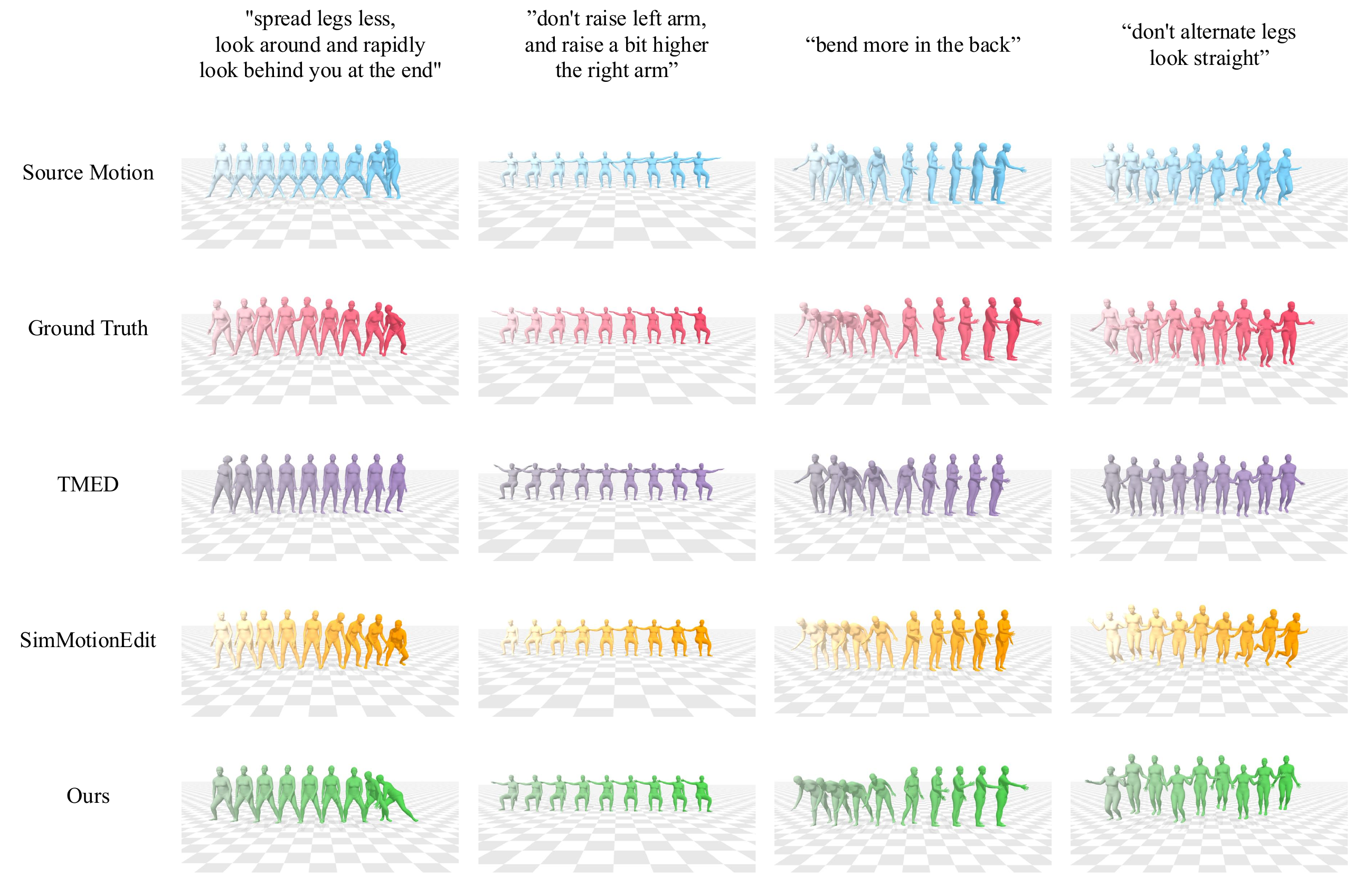}
      \caption{\textbf{Qualitative results.} We visualize the source motion, ground truth, and the edited motions from our method and competing methods, given a text instruction. To effectively illustrate the temporal progression, rendered meshes are translated to the right over time. For each motion, frame recency is encoded by saturation: lower saturation represents earlier frames, while higher saturation indicates more recent frames. Best viewed zoomed in.}
  \label{fig:qualitative_results}
\end{figure*}

\subsection{Experimental Setting}
\noindent\textbf{Dataset and baselines.}
We utilize the MotionFix dataset~\cite{MotionFix2024} for the training and evaluation of our model. This dataset, which is designed for text-based 3D motion editing, provides a total of 6,730 annotated samples. Each data point is structured as a triplet, containing a source motion, a target motion, and a corresponding text instruction. The dataset was constructed by first leveraging the TMR motion embedding space~\cite{TMR} to select semantically similar motion pairs from motion capture datasets. Following this, human annotators manually generated concise descriptions detailing the differences between the paired motions.

We designate TMED~\cite{MotionFix2024}, MotionReFit~\cite{DynBlend2025} and SimMotionEdit~\cite{SimMotionEdit2025} as our direct baselines, which are state-of-the-art (SOTA) methods trained on the same MotionFix dataset~\cite{MotionFix2024} capable of generating an edited motion solely from free-form text and a source motion. Additionally, following previous works, we also compare our proposed method against MDM~\cite{MDM2023} and MDM-BP~\cite{MotionFix2024}. MDM is a text-to-motion generation model, and MDM-BP is an adaptation of MDM for the text-based motion editing task that extracts body part labels via GPT~\cite{GPT4}.

\noindent\textbf{Evaluation metrics.}
To evaluate our method, we follow the motion-to-motion retrieval-based benchmark proposed in MotionFix~\cite{MotionFix2024}, which utilizes the pre-trained TMR~\cite{TMR} as a feature extractor. This benchmark assesses how well the edited motions are semantically aligned with both the text instructions and the source motions by checking how well the corresponding target motions and source motions are retrieved, respectively. Specifically, we compute the top-1, top-2, and top-3 retrieval accuracies (R@1, R@2, R@3), along with the average rank, within a batch size of 32. Furthermore, following the official implementation of SimMotionEdit, we also report these metrics on the test set for a more comprehensive evaluation. To assess the fidelity and diversity of the generated results, we additionally measure the Fréchet Inception Distance (FID) between our generated motions and the ground-truth target motions.

\noindent\textbf{Implementation details.}
We extract text embeddings from the text instructions for editing using a pre-trained CLIP~\cite{CLIP} (ViT-B/32) as previous works~\cite{MotionFix2024, SimMotionEdit2025}.
The joint-anchored transformer and the time-anchored transformer share an identical structure of 4 transformer encoder layers, 8 attention heads, and a 512-dimensional latent space. The diffusion transformer is deeper, built with 8 transformer encoder layers, 8 attention heads, and a 512 latent dimension. To merge the feature representations from the axis-anchored transformer modules, we designed a cross-axis fusion block, which consists of a multi-head cross-attention layer also configured with 8 attention heads and a 512 latent dimension. For training, we utilized the AdamW optimizer~\cite{AdamW} with a learning rate of $10^{-4}$ and a batch size of 64. The model was trained for 1,500 epochs, a process that required approximately 12 hours on a single NVIDIA RTX 4090 GPU. During the sampling phase, we adopted the inference configuration from prior studies~\cite{MotionFix2024, SimMotionEdit2025}. We employed the DDPM~\cite{DDPM} sampling process fixed at 300 diffusion steps, paired with a cosine noise scheduler. A two-way conditioning strategy was applied, setting the guidance scale to 2.0 for both the text and source motion inputs.

\subsection{Experimental Results}

\noindent\textbf{Quantitative results.}
As shown in Table~\ref{tab:quantitative_results}, the quantitative results demonstrate that the proposed method achieves state-of-the-art performance, outperforming all baselines across all retrieval-related metrics on both the batch and test set evaluations.
Specifically, our method achieves the highest Top-K retrieval accuracies (R@1, R@2, and R@3). The superior R@1 score is particularly significant, as it demonstrates that the motion edited by our model is most frequently the single closest match to the ground truth target motion. This indicates a high degree of precision in adhering to the editing instructions. Furthermore, our method obtains the best (lowest) average rank. This result signifies that, on average, the rank of the correct target motion is significantly closer to the top, highlighting the consistent accuracy and robustness of our model's generation capabilities compared to previous works.

In addition to semantic alignment with the text instructions and source motions, the fidelity and diversity of the generated motion are critical for overall quality. We evaluate this using the FID between the edited motions and the target motions. Our model achieves the lowest FID score and this result confirms that the motions generated by our approach are not only semantically accurate but also exhibit a distribution that is statistically closer to that of real motions, indicating superior realism and variety.

\begin{table*}[h!]
\centering
\begin{tabular}{lccccccccc}
\toprule
\multirow{2}{*}{Methods} & \multicolumn{4}{c}{Generated-to-Target (Batch)} & \multicolumn{4}{c}{Generated-to-Target (Test Set)} & \multirow{2}{*}{FID$\downarrow$}\\
\cmidrule(lr){2-5} \cmidrule(lr){6-9}
& R@1$\uparrow$ & R@2$\uparrow$ & R@3$\uparrow$ & AvgR$\downarrow$ & R@1$\uparrow$ & R@2$\uparrow$ & R@3$\uparrow$ & AvgR$\downarrow$ & \\
\midrule
Ground Truth & 100.0 & 100.0 & 100.0 & 1.00 & 64.36 & 88.75 & 95.56 & 1.74 & --\\
\midrule
MDM & 4.03 & 7.56 & 10.48 & 15.55 & 0.10 & 0.10 & 0.10 & -- & --\\
MDM-BP & 39.10 & 50.09 & 54.84 & 6.46 & 8.69 & 14.71 & 18.36 & 180.99 & --\\
TMED & 62.90 & 76.51 & 83.06 & 2.71 & 14.51 & 21.72 & 28.73 & 56.63 & 0.167\\
MotionReFit & 66.33 & 80.05 & 84.98 & 2.64 & - & - & - & - & - \\
SimMotionEdit & \underline{70.62} & \underline{82.92} & \underline{88.12} & \underline{2.38} & \underline{25.49} & \underline{39.33} & \underline{49.21} & \underline{23.49} & \underline{0.110}\\
\midrule
Ours & \textbf{74.38} & \textbf{88.54} & \textbf{92.08} & \textbf{1.92} & \textbf{29.45} & \textbf{45.26} & \textbf{54.55} & \textbf{16.42} & \textbf{0.097}\\
\bottomrule
\end{tabular}
\caption{\textbf{Quantitative results.} We compare the generated-to-target motion-to-motion retrieval performance of the proposed method with that of state-of-the-art text-based motion editing models. We report the top-1 (R@1), top-2 (R@2), and top-3 (R@3) retrieval accuracies along with the average rank (AvgR). To evaluate the fidelity and diversity of the generated motion, we additionally compute the FID between the edited motions and the target motions. Metrics are marked with $\uparrow$ (higher is better) or $\downarrow$ (lower is better). We highlight the \textbf{best} results in bold and the \underline{second-best} results with an underline.}
\label{tab:quantitative_results}
\end{table*}

\begin{table*}[h!]
\centering
\begin{tabular}{ccccccccccc}
\toprule
\multirow{2}{*}{Motion Sim.} & \multirow{2}{*}{Joint Delta} & \multicolumn{4}{c}{Generated-to-Target (Batch)} & \multicolumn{4}{c}{Generated-to-Target (Test Set)} & \multirow{2}{*}{FID$\downarrow$}\\
\cmidrule(lr){3-6} \cmidrule(lr){7-10}
& & R@1$\uparrow$ & R@2$\uparrow$ & R@3$\uparrow$ & AvgR$\downarrow$ & R@1$\uparrow$ & R@2$\uparrow$ & R@3$\uparrow$ & AvgR$\downarrow$ & \\
\midrule
\xmark & \xmark & 72.08 & 85.42 & 89.17 & 2.13 & \textbf{30.24} & 43.28 & 51.78 & 18.58 & 0.122\\
\xmark & L2 & 71.46 & 84.38 & 89.38 & 2.16 & 29.25 & 42.49 & 52.77 & 18.80 &  0.113 \\
\xmark & Soft-DTW & \underline{72.92} & 86.25 & 90.00 & 2.07 & 28.85 & 41.70 & 50.59 & 18.26  & \underline{0.108}\\
\cmark & \xmark & 71.04 & 85.42 & \underline{90.21}	& 2.03 &  29.45 & \underline{44.07} & \underline{54.35} & 17.98 & 0.143\\
\cmark & L2 & 71.04	& \underline{86.88}	& \underline{90.21}	& \underline{1.97} & 26.48 & 39.72	& 49.01	& \underline{17.54} & 0.113\\
\cmark & Soft-DTW & \textbf{74.38} & \textbf{88.54} & \textbf{92.08} & \textbf{1.92} & \underline{29.45} & \textbf{45.26} & \textbf{54.55} & \textbf{16.42} & \textbf{0.097}\\
\bottomrule
\end{tabular}
\caption{We conducted an ablation study to analyze the impact of auxiliary tasks on the text-based human motion editing performance. We evaluated the contributions of both the motion similarity prediction (Motion Sim.) task and the joint-wise distance prediction task (Joint Delta) by experimenting with whether each task was performed. For the joint-wise motion distance prediction task, we additionally conducted an experiment using the L2 distance as the distance metric, comparing it against Soft-DTW. Metrics are marked with $\uparrow$ (higher is better) or $\downarrow$ (lower is better). We highlight the \textbf{best} results in bold and the \underline{second-best} results with an underline.}
\label{tab:ablation_auxiliary}
\end{table*}

\noindent\textbf{Qualitative results.}
Recognizing that numerical metrics alone cannot fully capture the semantic coherence and perceptual quality of edited motions, we provide a qualitative comparison in Figure~\ref{fig:qualitative_results}. This figure visualizes the source motion, ground truth, and the results from our method and competing baselines for various text instructions.

As shown in the figure, TMED often struggles to accurately reflect the given text instructions, resulting in motions that are not semantically aligned with the edit. While both SimMotionEdit and our proposed method demonstrate significant improvements over TMED, our approach consistently generates motions that are more faithful to the textual commands.  

Furthermore, our method achieves stronger alignment with the source motion, retaining its original characteristics while precisely implementing the desired changes. We attribute this advantage to our model's architecture. We introduce an auxiliary task that trains the joint-anchored transformer to explicitly predict how each joint's motion should be altered (i.e., how much and in what way) according to the text instruction. The features from this transformer are then fused and provided as conditional input to the DiT. This mechanism enables our model to better discern which elements of the source motion must be preserved and which parts require editing.

\begin{table*}[h!]
\centering
\begin{tabular}{lccccccccc}
\toprule
\multirow{2}{*}{$\gamma$ for Soft-DTW} & \multicolumn{4}{c}{Generated-to-Target (Batch)} & \multicolumn{4}{c}{Generated-to-Target (Test Set)} & \multirow{2}{*}{FID$\downarrow$} \\
\cmidrule(lr){2-5} \cmidrule(lr){6-9}
& R@1$\uparrow$ & R@2$\uparrow$ & R@3$\uparrow$ & AvgR$\downarrow$ & R@1$\uparrow$ & R@2$\uparrow$ & R@3$\uparrow$ & AvgR$\downarrow$ & \\
\midrule
0.1 & 73.54	& 86.46	& 90.21	& 2.08 & 27.87 & \underline{42.69} & 51.19 & 20.13 & 0.107\\
0.5 & 72.08 & 85.00 & \underline{91.67}	& \underline{1.93} & 25.49 & 41.90 & 50.40 & \underline{17.20} & 0.112\\
1 & 73.54 & 86.25 & 89.38 & 2.00 & \underline{28.06} & 42.29 & \underline{52.57} & 17.81 & 0.115\\
5 & 73.75 & \underline{87.08} & 90.83 & 2.00 & 26.28 & 41.90 &  51.98 & 18.39 & \underline{0.106}\\
10 & \underline{74.38} & \textbf{88.54} & \textbf{92.08} & \textbf{1.92} & \textbf{29.45} & \textbf{45.26} & \textbf{54.55} & \textbf{16.42} & \textbf{0.097}\\
50 & \textbf{74.79} & 85.21 & 89.79 & 2.08 & 27.87 & \underline{42.69} & 51.78 & 19.21 & 0.108\\
\bottomrule
\end{tabular}
\caption{We analyze the model's performance by varying the $\gamma$ value for Soft-DTW from 0.1 to 50. Metrics are marked with $\uparrow$ (higher is better) or $\downarrow$ (lower is better). We highlight the \textbf{best} results in bold and the \underline{second-best} results with an underline.}
\label{tab:ablation_gamma}
\end{table*}

\begin{table*}[h!]
\centering
\begin{tabular}{lccccccccc}
\toprule
\multirow{2}{*}{Condition Feat.} & \multicolumn{4}{c}{Generated-to-Target (Batch)} & \multicolumn{4}{c}{Generated-to-Target (Test Set)} & \multirow{2}{*}{FID$\downarrow$} \\
\cmidrule(lr){2-5} \cmidrule(lr){6-9}
& R@1$\uparrow$ & R@2$\uparrow$ & R@3$\uparrow$ & AvgR$\downarrow$ & R@1$\uparrow$ & R@2$\uparrow$ & R@3$\uparrow$ & AvgR$\downarrow$ & \\
\midrule
Time-anchored & 70.62	& 82.92	& 87.92	& 2.28 & 24.11 & 37.35 & 46.05 & 22.93 & 0.111\\
Joint-anchored & \underline{72.08} & 84.58 & \underline{91.04} & 2.08 & \underline{27.67} & 40.91 & \underline{50.99} & 18.82 & 0.110\\
Mean & 70.42 & 83.54 & 89.58 & 2.15 & 26.48 & 40.71 & 48.62 & 20.79 & 0.114\\
Axis-fusion & 70.00 & \underline{84.79} & 90.00 & \underline{2.02} & 26.09 & \underline{41.50} & 48.81 & \underline{18.54} & \underline{0.108} \\
\midrule
Ours & \textbf{74.38} & \textbf{88.54} & \textbf{92.08} & \textbf{1.92} & \textbf{29.45} & \textbf{45.26} & \textbf{54.55} & \textbf{16.42} & \textbf{0.097}\\
\bottomrule
\end{tabular}
\caption{Ablation results on additional explicit text conditioning. Ours uses only the implicit fused feature $f_{\text{fusion}}$ without additional explicit text guidance. We highlight the \textbf{best} results in bold and the \underline{second-best} results with an underline.}
\label{tab:ablation_condition}
\end{table*}

\subsection{Ablation Study}
\noindent\textbf{Ablation study on the auxiliary tasks.} We conduct an ablation study to validate the effectiveness of our proposed auxiliary task and to analyze its individual impacts on performance. For the joint-wise motion distance prediction task, we investigate three variants: (1) not using the task (\xmark), (2) using L2 distance as the distance metric, and (3) using Soft-DTW as the distance metric on the rotation parameters.

As shown in Table~\ref{tab:ablation_auxiliary}, employing a simple L2 distance for the joint-wise motion distance prediction yields no significant performance improvement over the baselines that omit this task. We hypothesize that this is because the L2 loss is highly sensitive to temporal variations. For instance, if a motion has the similar semantic shape but is performed slightly faster or slower, the L2 distance can fluctuate dramatically. This volatility prevents the joint-anchored transformer from effectively learning to distinguish between temporal variations and genuine semantic changes required by the instruction. In contrast, using Soft-DTW leads to substantial performance gains across all retrieval metrics. This demonstrates that Soft-DTW, by being robust to temporal misalignments, provides a much more effective and stable learning signal. We believe this significant improvement arises because the Soft-DTW objective helps the model develop a more precise understanding of which joints to preserve from the source motion and which to modify according to the text.

\noindent\textbf{Ablation study on different $\gamma$ values.} 
We conducted an ablation study to observe the effect of the $\gamma$ hyperparameter used in the Soft-DTW calculation. In Soft-DTW, $\gamma$ serves as a smoothing parameter that controls the differentiability of the alignment operation; a smaller $\gamma$ approaches the non-differentiable standard DTW, while a larger $\gamma$ results in a smoother approximation. We evaluated the model's performance by varying $\gamma$ across six different values: {$0.1$, $0.5$, $1.0$, $5.0$, $10.0$, $50.0$}. As shown in Table~\ref{tab:ablation_gamma}, the model's performance is not overly sensitive to the choice of $\gamma$, showing robust results across a wide range of values.

\noindent\textbf{Effect of additional explicit text condition.} In our framework, the DiT model primarily receives text-related information implicitly through the fused feature $f_{\text{fusion}}$. Following previous work~\cite{SimMotionEdit2025}, we conducted an ablation study by introducing explicit text features from different modules as auxiliary conditioning variables to investigate whether providing explicit text guidance can further improve the model's performance. Specifically, we evaluated the effects of incorporating features from: (1) the time-anchored transformer, (2) the joint-anchored transformer, (3) the element-wise mean of the time-anchored and joint-anchored features, and (4) the axis-fusion transformer.

As shown in Table~\ref{tab:ablation_condition}, the results indicate that providing no explicit condition to the DiT achieves the best performance across all metrics. We hypothesize that this result is due to the information from the text instruction already being sufficiently encoded and infused into $f_\text{fusion}$, which serves as the primary input to the DiT. This suggests that an additional condition is redundant for the diffusion process.

\section{Conclusion}

In this work, we presented a 3D human motion editing framework capable of reasoning about the specific joints that drive intended motion updates. Our architecture employs axis-anchored transformers and a cross-axis fusion block to effectively decouple and integrate spatial and temporal cues from both source motion and text instructions. Furthermore, we introduced a joint-wise motion difference objective using Soft-DTW, which allows the model to discern which joints to modify versus preserve while maintaining robustness to temporal shifts. Extensive evaluations on the MotionFix dataset demonstrate significant improvements in semantic alignment and motion fidelity, validating the efficacy of our proposed architecture and auxiliary task.

\section*{Acknowledgments}
This work was supported by Institute of Information \& Communications Technology Planning \& Evaluation(IITP) grant funded by the Korea government(MSIT) (RS-2024-00439020, Developing Sustainable, Real-Time Generative AI for Multimodal Interaction, SW Starlab).

{
    \small
    \bibliographystyle{ieeenat_fullname}
    \bibliography{main}
}

\clearpage
\setcounter{page}{1}
\maketitlesupplementary
\setcounter{table}{0}
\renewcommand{\thetable}{\Alph{table}}

\setcounter{figure}{0}
\renewcommand{\thefigure}{\Alph{figure}}

\setcounter{equation}{0}
\renewcommand{\theequation}{\Alph{equation}}

\setcounter{section}{0}
\renewcommand{\thesection}{\Alph{section}}

\section{Generated-to-Source Motion Retrieval}

\begin{table}[h!]
\centering
\begin{tabular}{lcccc}
\toprule
\multirow{2}{*}{Methods} & \multicolumn{3}{c}{Generated-to-Source (Batch)}  & \multirow{2}{*}{FID$\downarrow$}\\
\cmidrule(lr){2-4}
& R@1$\uparrow$ & R@2$\uparrow$ & R@3$\uparrow$ & \\
\midrule
Ground Truth & 74.01 & 84.52 & 89.91 & - \\
\midrule
TMED & 71.77 & 84.07 & 89.52 &  0.167 \\
SimMotionEdit & 72.71 & 83.54 & 87.50 & 0.110\\
MotionReFit & \textbf{83.47} & \underline{90.42} & \underline{92.84} & -\\
Ours & \underline{78.96} & \textbf{91.04} & \textbf{93.33} & \textbf{0.097}\\
\bottomrule
\end{tabular}
\caption{We report generated-to-source motion retrieval results.}
\label{tab:supple_src}
\end{table}

To rigorously evaluate the capability of our model to maintain the structural and stylistic properties of the source motion $S$ while performing the instructed edit, we conducted generated-to-source motion retrieval experiments. This analysis is crucial for text-based motion editing tasks, as the generated output $M$ must exhibit a high degree of semantic consistency with $S$. Following the established TMR~\cite{TMR} retrieval protocol, we used the TMR feature space to check how well the generated motion retrieves its corresponding source motion. The results, shown in Table~\ref{tab:supple_src}, demonstrate that our method significantly outperforms the baselines, TMED~\cite{MotionFix2024} and SimMotionEdit~\cite{SimMotionEdit2025}. Furthermore, our approach achieves performance comparable to MotionReFit~\cite{DynBlend2025}, showing particularly strong results in R@2 and R@3. This superior performance indicates that our architecture—with its dedicated axis-anchored transformers and joint-aware supervision—ensures robust preservation of the source motion's semantic information while accurately implementing the desired textual modifications.

\begin{table*}[h]
\centering
\begin{tabular}{lccccccccc}
\toprule
\multirow{2}{*}{Methods} & \multicolumn{4}{c}{Generated-to-Target (Batch)} & \multicolumn{4}{c}{Generated-to-Target (Test Set)} & \multirow{2}{*}{FID$\downarrow$}\\
\cmidrule(lr){2-5} \cmidrule(lr){6-9}
& R@1$\uparrow$ & R@2$\uparrow$ & R@3$\uparrow$ & AvgR$\downarrow$ & R@1$\uparrow$ & R@2$\uparrow$ & R@3$\uparrow$ & AvgR$\downarrow$ & \\
\midrule
Motion Sim. before fusion & 70.00 & 84.79 & 90.00 & 2.00 & 26.09 & 41.50 & 48.81 & 18.54 & 0.122 \\
Larger fusion block & 72.92 & 86.88 & 90.42 & 2.02 & 29.45 & 46.05 & 55.14 & 16.65 & 0.103\\
Ours & \textbf{74.38} & \textbf{88.54} & \textbf{92.08} & \textbf{1.92} & \textbf{29.45} & \textbf{45.26} & \textbf{54.55} & \textbf{16.42} & \textbf{0.097}\\
\bottomrule
\end{tabular}
\caption{We report experiment results for different design choices of our framework.}
\label{tab:supple}
\end{table*}

\begin{table*}[h!]
\centering
% \resizebox{0.94\linewidth}{!}{
\begin{tabular}{lcccccc}
\toprule
\multirow{2}{*}{Methods} & \multicolumn{3}{c}{Generated-to-Target} & \multicolumn{3}{c}{Generated-to-Source} \\
\cmidrule(lr){2-4} \cmidrule(lr){5-7}
& R@1$\uparrow$ & R@2$\uparrow$ & R@3$\uparrow$ & R@1$\uparrow$ & R@2$\uparrow$ & R@3$\uparrow$ \\
\midrule
Ours w/ global dist. & 71.46 & 87.71 & 90.62 & 77.08 & 90.42 & 93.54\\
Ours w/ joint geodesic dist. & 72.50 & 85.42 & 89.79 & 78.33 & 89.58 & 93.33\\
Ours w/ classic DTW & 72.71 & 85.62 & 90.21 & 79.58 & 90.62 & 93.33\\
\midrule
Ours & 74.38 & 88.54 & 92.08 &  78.96 & 91.04 & 93.33\\
\bottomrule
\end{tabular}
% }
\caption{Ablation study on distance metrics for joint-wise supervision. We compare the impact of different distance objectives for training the joint-anchored transformer. We evaluate global frame-wise distance, joint-wise geodesic distance on the rotation manifold, and classic DTW against our proposed Soft-DTW.}
\label{tab:supple_distance}
\end{table*}

\begin{figure*}[t!]
    \centering
    \includegraphics[width=1.0\linewidth]{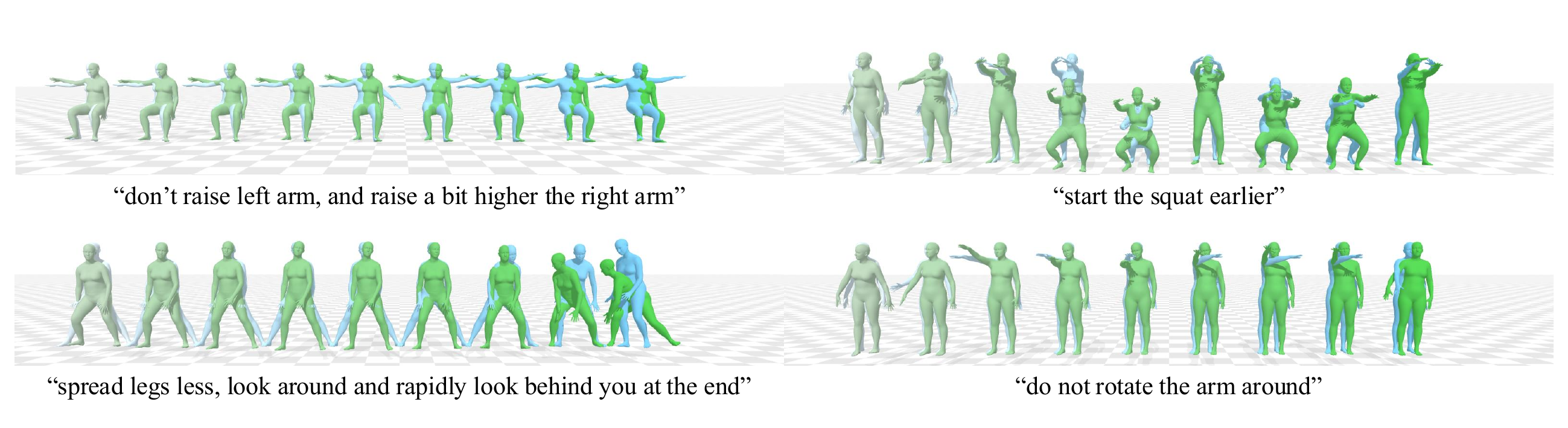}
            \caption{We visualized source motions (blue) and motions edited with our method (green) with overlay.}
    \label{fig:supple_overlay}
\end{figure*}

\section{Experiments on Different Design Choices}
We conducted experiments on various design choices within our framework.

\subsection{Motion Similarity Prediction Before Fusion}

We conducted an ablation study to investigate the optimal placement of the auxiliary motion similarity prediction task, originally introduced in SimMotionEdit~\cite{SimMotionEdit2025}. In our final proposed architecture, the task is applied to the features after passing through the cross-axis fusion block. For this ablation, we shifted the motion similarity prediction head to operate on the intermediate frame-wise feature $h_{\text{time}}$, specifically, the output of the time-anchored transformer just before it enters the cross-axis fusion block.

As shown in Table~\ref{tab:supple}, the performance achieved by applying the motion similarity prediction task to the intermediate $h_{\text{time}}$ feature is consistently lower than our proposed method.
This outcome validates our architectural design: the joint-anchored transformer is explicitly trained by the Soft-DTW auxiliary objective ($\mathcal{L}_{\text{aux}}$) to understand which joints to modify and which to preserve. By performing the motion similarity prediction task after the fusion block, the prediction head leverages the rich, contextualized representation ($h_{\text{fusion}}$) that the cross-axis fusion block integrates from both joint-aware and temporally-aware conditioning. This results in superior performance, as the fused feature provides a more robust basis for predicting the overall frame-wise motion similarity.

\subsection{Larger Cross-Axis Fusion Block}

We further analyzed the impact of the design of the cross-axis fusion block on model performance. In our main design, we employ a single multi-head cross-attention block to integrate the time-anchored feature ($h_{\text{time}}$ as Query) and the joint-anchored feature ($h_{\text{joint}}$ as Key and Value). For this ablation, we constructed a significantly larger fusion block consisting of an alternating sequence of 4 temporal self-attention blocks and 4 cross-attention blocks.

The results presented in Table~\ref{tab:supple} show that the performance with the larger, more complex fusion block was consistently inferior compared to our optimized single cross-attention layer design.
We hypothesize that this degradation is due to two primary factors. First, the substantial increase in the number of parameters within the fusion pathway may lead to overfitting on the MotionFix dataset, which is moderate in size for complex models. Second, the repetitive self-attention layers within the fusion block may unnecessarily re-aggregate the features along the temporal axis. This repetition potentially disrupts the explicit disentanglement between the joint-wise and time-wise information that our axis-anchored transformers were designed to establish, thereby diluting the precise and targeted conditioning signal needed for fine-grained motion editing. The simpler, single cross-attention block proves sufficient to perform the required feature integration without introducing undue complexity or feature corruption.

\section{Analysis of Distance Metrics for Motion Difference Prediction}
We conducted several experiments to justify the design of our auxiliary task in \Cref{tab:supple_distance}. First, the lower performance of the global distance metric compared to our approach suggests that holistic motion alignment is insufficient for the encoder to reason about joint-specific signals. Regarding the metric formulation, we observed that the joint geodesic distance yields suboptimal results because it collapses multi-dimensional rotation differences into a single scalar, which restricts the information available for supervision. In contrast, our channel-wise prediction provides higher degrees of freedom through finer-grained signals, enabling the encoder to capture the complex dynamics of 6D rotations more effectively. Finally, our method outperforms Classic DTW because the soft-min operator is more robust to high-frequency motion noise than rigid alignment, thereby providing a more stable and differentiable learning signal for the transformer.

\section{More Qualitative Results}
We provide an overlay visualization of source motions and motions edited with our method in \Cref{fig:supple_overlay}, superimposing edited motions onto the original source sequences.

\section{Future Work}

Looking ahead, we aim to extend our framework by leveraging the rich prior knowledge of emerging large-scale text-to-motion foundation models. As motion generation enters the era of large-scale pre-training with massive datasets and significantly increased parameter counts, we intend to investigate how our proposed axis-anchored conditioning and joint-wise auxiliary task can be effectively utilized to fine-tune these models for the motion editing task. We believe that integrating our structural and joint-aware supervision with the broad motion priors of foundation models will further enhance the precision and diversity of text-based editing.

Furthermore, we aim to further validate the generalizability of our framework across a wider variety of datasets, such as the STANCE benchmark~\cite{DynBlend2025}. While we initially explored an evaluation on the STANCE benchmark, a fair comparison was not feasible at this stage due to the current lack of a publicly available implementation for the full training and evaluation protocol. We look forward to conducting these experiments once the standardized pipeline is released by the authors.

% WARNING: do not forget to delete the supplementary pages from your submission 

\end{document}